\documentclass[a4paper,fleqn]{cas-dc}

\usepackage[authoryear,longnamesfirst]{natbib}

\usepackage[T1]{fontenc}
\usepackage{graphicx}
\usepackage{listings}
\usepackage{varwidth}
\usepackage[dvipsnames]{xcolor}
\usepackage{url}
\usepackage{subcaption}
\usepackage{adjustbox}
\usepackage{comment}
\usepackage{multirow}
\usepackage{booktabs}
\usepackage{enumitem}
\usepackage[most]{tcolorbox}
\usepackage{xurl}
\usepackage{hyperref}
\usepackage{etoolbox}

\definecolor{roleblue}{RGB}{0,102,204}
\definecolor{systemgreen}{RGB}{102,170,0}
\definecolor{contextred}{RGB}{230,60,60}
\definecolor{feworange}{RGB}{245,155,0}
\newcommand{\fewshot}[1]{\colorbox{feworange!70}{#1}}
\definecolor{yelloworange}{RGB}{255,200,80}

\setlist[itemize]{leftmargin=*, itemsep=0.6mm, topsep=0.6mm}

\begin{document}
\let\WriteBookmarks\relax
\def\floatpagepagefraction{1}
\def\textpagefraction{.001}


\title[mode=title]{LLM-Orchestrated Conformance Checking in Stroke Care Without Computer-Interpretable Guidelines}

    
\author[1]{Giorgio Leonardi}[type=author,
                        auid=001,bioid=1,
                        orcid=0000-0002-9533-9722]
\ead{giorgio.leonardi@uniupo.it}

\author[1]{Stefania Montani}[type=author,
                        auid=002,bioid=2,
                        orcid=0000-0002-5992-6735]
\ead{stefania.montani@uniupo.it}
\cormark[1]

\author[1]{Manuel Striani}[type=author,
                        auid=003,bioid=3,
                        orcid=0000-0002-7600-576X]
\ead{manuel.striani@uniupo.it}

\author[2]{Alessandro Canessa}
\ead{alessandro.canessa@ospedale.al.it}

\author[2]{Delfina Ferrandi}
\ead{dferrandi@ospedale.al.it}

\affiliation[1]{organization={Computer Science Institute, DiSIT, University of Piemonte Orientale},
    addressline={Viale Teresa Michel, 11},
    city={Alessandria},
    postcode={15121},
    country={Italy}}

\affiliation[2]{organization={Integrated Laboratory of AI and Medical Informatics, DAIRI, SS. Antonio e Biagio e Cesare Arrigo Hospital},
    city={Alessandria},
    country={Italy}}
    \cortext[cor1]{Corresponding author}

\begin{abstract}
\textbf{Objective} Conformance checking in healthcare seeks to assess whether patient care pathways adhere to clinical guidelines. However, its practical application often depends on the availability of formal, machine-interpretable representations of guidelines, such as Computer-Interpretable Guidelines (CIGs), which are seldom available in real-world clinical settings.

\textbf{Methods} This work introduces a modular framework based on the orchestration of Large Language Models (LLMs) to support medical conformance checking directly from unstructured clinical and guideline texts, without requiring predefined CIGs. The proposed architecture integrates multiple LLMs and supporting components to extract patient traces from clinical discharge letters, identify normative rules from textual clinical guidelines, translate these rules into executable scripts, and compute a Trace Conformance Indicator to quantify compliance within the event log.

\textbf{Results} The framework was implemented and evaluated in the stroke care domain at the neurological ward of Alessandria Hospital. Hundreds of patient traces were automatically extracted from hospital data and assessed against 50 rules derived from the reference guideline. The analysis showed that more than 86\% of the available traces were conformant.

\textbf{Conclusion} The results demonstrate the feasibility of using orchestrated LLMs for practical healthcare conformance analysis. At the same time, the study provides evidence of a high level of adherence to stroke care guidelines at Alessandria Hospital.

\end{abstract}


\maketitle

\section{Introduction}

Conformance checking is a fundamental activity in Process Mining (PM) \citep{aalst:book:16} that assesses whether an organization’s observed behavior complies with its prescribed processes. Observed behavior is typically captured in an event log, which consists of a collection of process traces \citep{Reichert:2012}. Each trace records the sequence of activities carried out during routine operations, along with their timestamps. In the healthcare domain, a trace represents the ordered set of activities performed for a patient throughout a hospital stay. The behavior prescribed in this context is typically defined by a clinical guideline.

Conformance checking techniques rely on two main inputs: the event log and a formal, machine-readable specification of the normative process. This specification is commonly provided as a graphical control-flow model expressed in a dedicated modeling language, which can be converted into the well-known Petri Net formalism \citep{DBLP:conf/ac/2003pn}. In healthcare settings, however, such formal representations are rarely available, as clinical guidelines are usually documented in extensive and complex natural-language texts. Although methods for developing Computer-Interpretable Guidelines (CIGs) have been proposed \citep{ClercqKH08}, they require significant effort in terms of knowledge elicitation and modeling.

In this paper, we introduce a 
framework that orchestrates different Large Language Models (LLMs) to support medical conformance checking by validating patient traces directly on a textual guideline, without the need for an explicit CIG.

Our approach complies with a recent trend in  Artificial Intelligence (AI) applications, where we are assisting to a shift away from reliance on ever-larger standalone models toward compound systems, in which multiple components are deliberately orchestrated to solve complex tasks \citep{bair2024compound}. These systems can integrate LLMs with auxiliary modules such as retrieval mechanisms, symbolic solvers, execution engines, and verification steps, and/or chain different LLMs to complete the task. Empirical evidence from state-of-the-art systems demonstrates that such structured compositions can outperform monolithic models by enabling multi-step reasoning, solution filtering, and interaction with external knowledge sources. In practice, techniques such as chained reasoning pipelines have become common, underscoring that system-level orchestration design choices often yield larger performance gains than additional model scaling alone (see, e.g., \citep{wrike2025aiworkflow,palani2025orchid}).


Within this line of investigation, our contribution consists in the design, implementation and testing of a modular architecture, where:
\begin{enumerate}
\item  a first LLM extracts patient traces from unstructured clinical discharge letters, and exports them in XES format\footnote{\url{https://www.xes-standard.org/}} for PM analyses; 
\item  a second LLM extracts normative behaviors from a textual guideline in the form of rules;
\item a third LLM filters out the un-useful rules, i.e., those that cannot be applied to the given event log (typically due to data unavailability);
\item a fourth LLM formalizes the rules into executable Python scripts; 
\item a fifth LLM refines the formalized rules by eliminating noise and artifacts, as well as by correcting bugs;
\item a Python module ultimately computes the Trace Conformance Indicator ($\mathcal{TCI}$), a metric that measures the proportion of event log traces satisfying each formalized rule. This enables clinicians and healthcare administrators to rapidly detect conformance violations or operational issues in real-world event logs.
\end{enumerate}

The overall architecture has been extensively tested in the domain of stroke care, where we were able to extract 463 patient traces from real discharge letters collected at Alessandria Hospital, and to assess their conformance towards one of the  most recently published Italian stroke clinical guidelines \citep{GL}. 
Details are provided in the following sections. In particular, section~\ref{related} reviews related work; section~\ref{workflowarchitecture} technically describes the architecture; section~\ref{experiments} presents experimental results, while section~\ref{conclu} is devoted to conclusions.

\section{Related Work}
\label{related}

In this section, we present related work, focusing first on classical conformance checking approaches, then on the use of LLMs in Process Mining, and finally on LLM orchestration.

\subsection{Conformance checking}

In the Process Mining literature (see, for example, \citep{DunzerSMB19}), conformance checking techniques are commonly classified into three broad categories: \emph{log replay methods, trace alignment methods, and constraint-based approaches}.

\emph{Log replay methods} simulate the execution of each trace from the event log on the prescribed process model. When the normative model is represented as a Petri Net, token-based replay techniques are typically employed \citep{rozinat:08}, allowing the generation of additional tokens to overcome deadlocks and complete the execution; conformance is then assessed based on the number of missing or remaining tokens. In more general settings, where the model is not a Petri Net, conformance can be quantified using fitness metrics \citep{Leemans:2018}, defined as the proportion of traces that can be successfully replayed with respect to the total number of traces.

\emph{Trace alignment methods} identify deviations at the level of individual activities by detecting asynchronous moves, that is, steps where the trace and the model cannot advance simultaneously. Each deviation is associated with a cost, and an optimal alignment minimizing the overall cost is computed \citep{AdriansyahMCDA12}. While this approach yields more fine-grained and precise diagnostics than log replay, it is computationally more demanding, as the number of possible alignments increases rapidly with the size of the event log.

\emph{Constraint-based conformance checking approaches} (e.g., \citep{Borrego014}) represent process behavior in terms of declarative rules and assess whether traces satisfy these constraints. The method proposed in this work falls within this category, with the key novelty that, to the best of our knowledge, it is the first to leverage an LLM to automatically extract such rules from a normative specification expressed in free natural language, within a larger orchestrated architecture for conformance checking in medicine.

\subsection{LLMs in Process Mining}

Large Language Models (LLMs), including BERT \citep{devlin-etal-2019-bert} and Google's Gemini \citep{gemini2}, have significantly expanded the state of the art in Natural Language Processing (NLP) \citep{DBLP:journals/mta/KhuranaKKS23}, opening up new possibilities for applications in Process Mining (PM). 

An overview of recent research at the intersection of LLMs and PM is provided in the survey by \citep{berti2024evaluating}.

According to the survey, on the one hand, LLMs have been used to automatically produce natural language text and explanations from event logs and process models \citep{DBLP:conf/bpm/Berti0A23,kourani2024}, as well as from visual artifacts such as dotted charts \citep{Qafari}. To improve process interpretability, researchers have adopted techniques including abstraction \citep{DBLP:conf/bpm/Berti0A23} and sophisticated prompt engineering strategies \citep{kourani2024}. LLMs have also been leveraged to support natural language querying of process data, for example to identify performance bottlenecks or detect fairness-related issues \citep{Qafari}.

On the other hand, LLMs have been used to extract process elements and/or process models from textual documents, by leveraging prompt engineering, error correction, and code generation mechanisms \citep{Kourani_2024}. The work in \citep{Susaiyah} adopts a LLM-based approach to extract event logs from a human-annotated dataset of nurse notes in the MIMICS-III dataset. The work in \citep{grohs2023large} explores the use of LLMs to derive both imperative and declarative process models from text and to evaluate process suitability for robotic process automation, achieving improvements over earlier NLP-based techniques that were confined to narrower tasks. Interactive refinement of the generated models is also supported in \citep{Kourani_2024}.
Beyond model extraction, LLMs have been applied to generate SQL queries for filtering and analyzing event logs stored in relational databases \citep{jessen2023chitchat}.

In contrast to prior work on declarative PM (see, e.g., \citep{grohs2023large}), our objective is more focused: we do not seek to discover a complete declarative process model, and the required knowledge acquisition effort is correspondingly lower (while, at the same time, we also work on event log extraction from natural text, as in \citep{Susaiyah}).  Furthermore, among the existing literature, only \citep{DBLP:conf/bpm/Berti0A23} explicitly addresses LLM-assisted conformance checking; however, its results on medical datasets are limited, and it does not propose a method for deriving rules directly from domain knowledge.

\subsection{LLM orchestration}

A growing number of AI architectures are being proposed in the literature, that orchestrate different LLMs, as well as other components, to complete a reasoning task, rather than developing excessively complex single models. Focusing on healthcare applications, LLM‑Synergy \citep{yang2025llmensemble}, for instance, is an ensemble learning framework that integrates outputs from multiple LLMs to improve performance on medical question‑answering datasets. Always in medical question-answering, HealthBranches \citep{cosentino} first extracts structured clinical knowledge from medical literature, and then exploits it to score different LLMs' answers through LLM-as-a-Judge evaluation.

Orchestration can also take place in an agentic fashion: agentic AI denotes autonomous AI systems capable of defining objectives, planning and reasoning about actions, and carrying out complex tasks with little human intervention. Unlike traditional command–response models, these systems operate as independent agents that adapt to changing environments and learn from experience to accomplish their goals. Agentic AI typically integrates LLMs with additional tools and algorithms to coordinate multi-step workflows, enabling proactive problem solving.
In healthcare, Rodella et al. \citep{rodella2025agentic}, for instance, describe an agentic AI workflow that integrates multiple specialized LLMs into a coordinated pipeline to automate the design of simulation scenarios. The system decomposes complex tasks (such as clinical narrative generation, diagnostic data creation, and educational debriefing) into sub-tasks handled by different AI agents, and these are orchestrated using advanced techniques like prompt chaining, parallelization, retrieval augmentation, and iterative refinement — demonstrating how orchestrated AI workflows can significantly improve efficiency and usability in medical applications.

Our  approach is not (yet) designed as agentic AI (this line of research will possibly be considered in the future). However, it still addresses the goal of reducing the systematic biases that may emerge when fully relying on a single model, enhancing robustness and generalizability. 
To the best of our knowledge, it represents the first example of a LLM orchestration architecture able to support medical process conformance checking, drastically reducing the knowledge acquisition bottleneck from natural language, working at the same time both on the formalization of the event log and on the formalization of the normative rules.  

\section{LLM-assisted Conformance Checking}
\label{workflowarchitecture}

\begin{figure*}
\begin{adjustbox}{center}
    \centering
    \includegraphics[width=\textwidth]{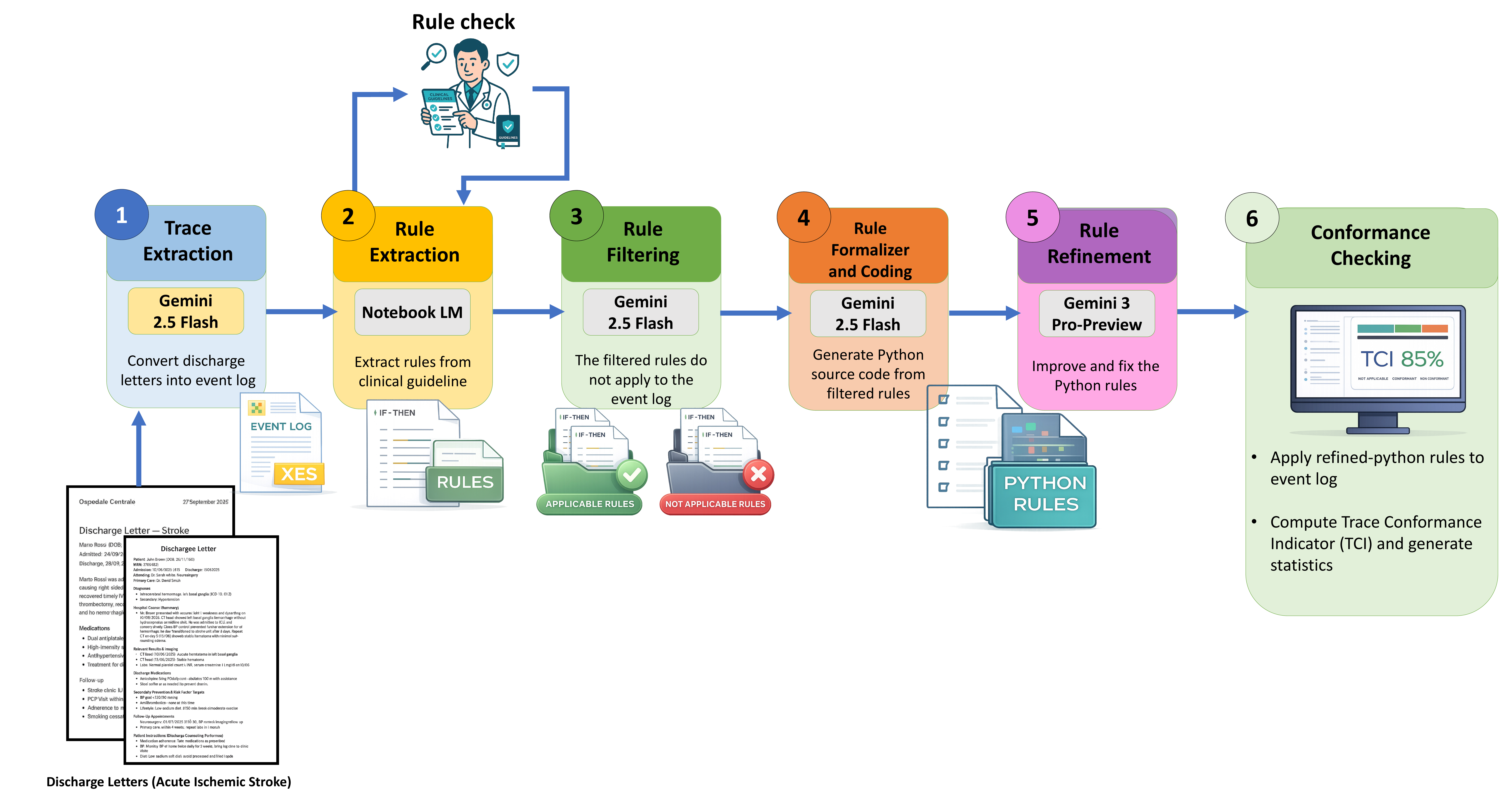}
    \end{adjustbox}
    \caption{Orchestrated LLM pipeline for guideline-based conformance checking}
    \label{LLMs_architecture}
\end{figure*}

In our architecture, illustrated in Figure \ref{LLMs_architecture}, we orchestrate four LLMs, as well as a final Python module, to realize a pipeline, which accomplishes the following steps:

\begin{enumerate}
    \item \emph{Trace Extraction}: by means of a first LLM (currently, Gemini 2.5 Flash \citep{gemini2}), we process discharge letters, and convert them into an event log, formalized as a file of XES traces;
    \item \emph{Rule Extraction}: by means of a second LLM (currently, NotebookLM \citep{notebook}), we extract a set of semi-structured rules (in the IF-THEN form) from a textual clinical guideline; in order to solve possible issues, we implement a human-in-the-loop strategy, where the physician identifies incorrect or incomplete rules with respect to medical knowledge, and (possibly with the help of a technical expert) refines the prompt. The LLM is thus properly instructed to correct the original rule set, which is evaluated again, until the result is satisfactory;
   
     \item \emph{Rule Filtering}: by means of a third LLM (currently, Gemini 2.5 Flash \citep{gemini2}), we filter out those rules that cannot be applied to the available event log. In fact, some rules refer to guideline recommendations that are not meant to be applied in the hospitalization phase (e.g., they refer to prevention), or simply require patient data that were not routinely collected in our dataset (and thus they cannot be checked). Separating the rule filtering phase (which requires the provision of the event log to inform the LLM on what data are available) from the previous rule extraction phase allows us to obtain an initial rule set which is not biased by the current event log content, but only depends on the guideline;
       \item \emph{Rule Coding}: by means of a fourth LLM (currently, still Gemini 2.5 Flash \citep{gemini2}), we convert the filtered rules into executable Python scripts; when applied to a trace, the script will provide one of the following answers: \emph{not applicable, conformant, not conformant}. A rule may be \emph{not applicable} to a given trace because, e.g., it refers to a procedure that was not applied to that specific patient; \emph{not conformant}, instead, means that the procedure was applied, but in a non-correct way with respect to the guideline directives (e.g., too late);
    \item \emph{Rule Refinement}: by means of a fifth LLM (currently, Gemini 3 Pro-Preview \citep{gemini3}) we  improve the Python rules, by fixing possible coding bugs, and eliminating redundancy, noise and artifacts; indeed, Gemini 3 Pro-Preview exceeds the capacity of the other LLM models, and is thus suitable to identify intrinsically flawed items;
    \item \emph{Conformance Checking}: by means of a last Python module, we apply the refined rules to the event log, and compute a metric that we termed Trace Conformance Indicator ($\mathcal{TCI}$), which provides the percentage of conformant traces for each applicable rule.
\end{enumerate}


In order to extract traces from the discharge letters, we adopt a \emph{few-shot prompting} technique \citep{liu2023pretrain,prompt}, where, along with the task description, we provide examples of trace format and content (specifically XES tags, activity reference lists, activity names).


\begin{figure*}
  \centering
  \begin{adjustbox}{max totalsize={\textwidth}{0.97\textheight},center}
    \begin{minipage}{0.98\textwidth}
      \footnotesize
      \sloppy

      \begin{tcolorbox}[
        enhanced,
        colback=white,
        colframe=roleblue,
        boxrule=2pt,
        arc=1.6mm,
        left=2.5mm, right=2.5mm, top=1.6mm, bottom=1.6mm
      ]
      \textbf{You are an expert in clinical process mining.}\\
      Your task is to analyze a hospital discharge letter and generate a process trace in XES format,
      following the official standard as rigorously as possible.
      \end{tcolorbox}

      \vspace{0.8mm}

      \begin{tcolorbox}[
        colback=white,
        colframe=systemgreen,
        boxrule=2pt,
        arc=1.6mm,
        left=2.3mm, right=2.3mm, top=1.6mm, bottom=1.6mm
      ]
      \textbf{Follow these rules strictly and without exceptions:}

      \vspace{0.8mm}

      \textbf{1. XES STRUCTURE RULES*:}
      \begin{itemize}
        \item The output must be a complete and valid XML file that starts with
        \texttt{<?xml version="1.0" encoding="UTF-8"?>}.
        \item The root tag must include the necessary extensions:
        \begin{itemize}
          \item[-] \texttt{<log xes.version="1.0" xmlns="http://www.xes-standard.org/">}
          \item[-] \texttt{<extension name="Concept" prefix="concept" uri="http://www.xes-standard.org/concept.xesext"/>}
          \item[-] \texttt{<extension name="Time" prefix="time" uri="http://www.xes-standard.org/time.xesext"/>}
        \end{itemize}
        \item All events must be in \textbf{strict chronological order} and contained inside a \texttt{<trace>} tag.
        \item Each event (\texttt{<event>}) must contain EXACTLY the following tags and \texttt{key} attributes:

        \begin{tcolorbox}[
          colback=yelloworange,
          colframe=yelloworange,
          boxrule=2pt,
          arc=1.2mm,
          left=2mm, right=2mm, top=1.2mm, bottom=1.2mm
        ]
        \begin{itemize}
          \item[-] For the timestamp: \texttt{<date key="time:timestamp" value="..."/>}
          \item[-] For the activity name: \texttt{<string key="concept:name" value="..."/>}
          \item[-] For details: \texttt{<string key="note" value="..."/>}
        \end{itemize}
        \end{tcolorbox}
      \end{itemize}

      \vspace{0.8mm}

      \textbf{2. RULES FOR ACTIVITY NAMES (\texttt{concept:name})*:}
      \begin{itemize}
        \item \textbf{LANGUAGE:} The \texttt{value} attribute must ALWAYS be in \textbf{Italian}*.
        \item \textbf{GOLDEN RULE OF STANDARDIZATION (MANDATORY)*:}\\
        For the value of \texttt{<string key="concept:name" value="..."/>}, you MUST use ONLY the STANDARD NAMES
        present in the reference list below. Analyze the letter text, identify an event, find its variant in the
        reference list, and use the corresponding STANDARD name. You are not allowed to invent new standard names if a mapping exists. The reference list has absolute priority.
        \item \textbf{RULE FOR ACTIVITIES NOT PRESENT IN THE LIST*:}\\
        If an activity in the text does not match any variant in the reference list, create a new standard activity
        name. Use the structure of STANDARD names in the list (e.g., \fewshot{``Therapy Start...''},
        \fewshot{``Neurological Assessment...''}, \fewshot{``CT Scan Execution...''}) and create a new name that is:
        \begin{itemize}
          \item[-] \textbf{Concise and Clear*:} \texttt{`Noun + Adjective'} or \texttt{`Verb + Object'}
          (e.g., \fewshot{``Cardiology Visit''}).
          \item[-] \textbf{Consistent*:} Same terminology and style as the list.
          \item[-] \textbf{In Italian and with Capital Initials*.}
        \end{itemize}
      \end{itemize}
      \end{tcolorbox}

      \vspace{0.8mm}

      \begin{tcolorbox}[
        colback=white,
        colframe=contextred,
        boxrule=2pt,
        arc=1.3mm,
        left=2.3mm, right=2.3mm, top=1.2mm, bottom=1.2mm
      ]
      \texttt{--- START OF ACTIVITY REFERENCE LIST ---}\\
      \texttt{\{standardization\_rules\}}\\
      \texttt{--- END OF ACTIVITY REFERENCE LIST ---}
      \end{tcolorbox}

      \vspace{0.8mm}

      \begin{tcolorbox}[
        colback=white,
        colframe=systemgreen,
        boxrule=2pt,
        arc=1.6mm,
        left=2.3mm, right=2.3mm, top=1.2mm, bottom=1.2mm
      ]
      \textbf{FINAL REVIEW*:} Before generating the output, reread every tag
      \texttt{<string key="concept:name" value="...">} and ensure its value complies with the rules.
      \end{tcolorbox}

      \vspace{0.8mm}

      \begin{tcolorbox}[
        colback=white,
        colframe=contextred,
        boxrule=2pt,
        arc=1.3mm,
        left=2.3mm, right=2.3mm, top=1.2mm, bottom=1.2mm
      ]
      Generate the XES trace for the following text:\\
      \texttt{--- START OF FILE ---}\\
      \texttt{\{letter\_content\}}\\
      \texttt{--- END OF FILE ---}
      \end{tcolorbox}

      \vspace{2mm}

      \textsl{\textbf{Legend}}

      \vspace{1mm}

      \begin{tcolorbox}[
        colback=white,
        colframe=black!20,
        boxrule=1.1pt,
        arc=1.2mm,
        left=2mm, right=2mm, top=1.2mm, bottom=1.2mm
      ]
      {\color{roleblue}\rule{10mm}{3pt}}\; Role prompting\\
      {\color{systemgreen}\rule{10mm}{3pt}}\; System prompting\\
      {\color{contextred}\rule{10mm}{3pt}}\; Contextual prompting\\
      {\color{yelloworange}\rule{10mm}{3pt}}\; Few-shot prompting\\[0.6mm]
      \textit{\textbf{*} Implicit zero-shot prompting}
      \end{tcolorbox}

    \end{minipage}
  \end{adjustbox}
  \caption{Prompt template with color-coded prompting strategies. The original prompt (in Italian) has been translated into English for the sake of paper clarity.}
  \label{fig:prompt-strategies}
\end{figure*}

Figure \ref{fig:prompt-strategies} illustrates the adopted prompt, highlighting the different strategies \citep{liu2023pretrain,prompt} we use. Specifically, we resort to:
\begin{itemize}
    \item \emph{role prompting}: we assert that the LLM is a medical PM expert, to force the model (which is a frozen one) to use already available domain knowledge, and an approach more focused to the task;
    \item \emph{contextual prompting}: to build context, we provide the discharge letter, and an activity synonym reference list (i.e., a vocabulary). In particular, such a reference list had been learned by the LLM itself in a previous step of the work, where all the discharge letters were provided to the model; the list was then validated by medical experts. In the future we will consider to integrate a standard vocabulary, such as SNOMED-CT\footnote{\url{https://www.snomed.org/}}, along the lines described in \citep{mie2026}, in a Retrieval Augmented Generation (RAG) \citep{RAG-nuovo,prompt} fashion;
    \item \emph{system prompting}: we establish some general rules that the model has to follow when completing the task, such as respecting the XES syntax, use reference list terms in the traces, use Italian.
\end{itemize}


A similar strategy is used for prompting in the other phases of the work (i.e., rule extraction, rule filtering, rule coding and refinement). In particular, contextual prompting for rule filtering, as already observed, requires the provision of the event log, while rule coding  requires a (random) example trace in order to infer its structure. 


%





As regards the choice of the LLMs, it was basically dictated by the availability of Google tools within an educational agreement active at our university. In particular, we also tested Gemini for rule extraction, but it only provided a small rule set (16 rules), leading us to prefer NotebookLM for this task. NotebookLM, which extracted 161 rules (see section \ref{experiments}), also proved more accurate with respect to GPT-5 \textsl{Thinking}, which we tested in our past work \citep{hydra} (where we adopted a different event log), that only provided 28 rules. Possible further tests will be considered in the future.








\section{Experiments}
\label{experiments}

 Stroke is a serious medical emergency that occurs when blood supply to the brain is disrupted, leading to the death of brain cells. This disruption can result from ischemia—caused by a blood clot or embolus that deprives brain tissue of oxygen and glucose—or from hemorrhage. During the acute stage, stroke poses a significant threat to life, and survivors frequently experience severe complications that may result in long-term or permanent disability. Stroke is the primary cause of disability among adults in the United States and Europe and is the second leading cause of death worldwide.

In our experiments, we considered 463 discharge letters of real stroke patients (properly anonymized), cared at the neurology ward of Alessandria Hospital (Italy), over the years 2022-2024\footnote{Patient data cannot be made available, due to privacy reasons. Instead, our code is available at \url{https://osf.io/qv36n/overview?view_only=7a5f968ecbf143b09a407e9205546b9e}}. The aim of the study was that of translating the letters into process traces, in order to automatically check the conformance of the implemented activities  with respect to the Italian reference guideline, in a quality assessment perspective.


Gemini 2.5 Flash extracted an event log of 463 items, composed of 47 activities on average, corresponding to a mean hospitalization duration of 10 days. In the current work, a random subsample (20\%) of the extracted traces was checked and validated by medical experts.  The physicians verified the coherence between the discharge letter information and the trace, the correctness of activity names (taking into account the synonym dictionary), the correctness of the timestamps (when available) and/or  of the temporal ordering of the activities. 

All the checked traces were judged as correct and usable, and a human-in-the-loop iteration, asking the LLM to refine the output, was not implemented. Such an iterative perspective can however be considered in the future, along the lines sketched in \citep{mie2026}. 

NotebookLM then extracted 161 rules from the Italian guideline \citep{GL}, organized into 9 categories: (1) diagnosis and initial imaging (20 rules); (2) acute phase management of ischemic stroke (19 rules); (3) acute phase management of hemorrhagic stroke (16 rules); (4) monitoring and complications (20 rules); (5) primary prevention (24 rules); (6) secondary prevention with antithrombotic therapy (15 rules); (7) secondary prevention with surgery (12 rules); (8) rehabilitation (19 rules); (9) special populations (16 rules).

Our medical collaborators  checked  the rules extracted by the LLM, focusing especially on the first 4 categories, which are connected with the work of the neurological ward. 

Rules are expressed as simple IF-THEN statements, and were easy and fast to check by the human experts. The rules were deemed both semantically accurate —  reflecting the medical knowledge in the guideline — and clinically relevant, as their application is necessary for delivering correct and timely patient care. Overall, a good synthesis capability (with respect to the original, quite verbose textual guideline) was demonstrated by the LLM, with the exception of the secondary prevention rules, which did not differ much from synthetic guideline versions that are usually available to physicians for quick reference. It is however worth noting that secondary prevention rules do not relate to acute phase hospitalization, and were thus not considered in the further steps of our experiments.

On a positive note,  "less crucial" information (e.g.,  summaries of clinical trial outcomes, that the guideline reports to support recommendations) was properly left out, and the extracted rules were free from hallucinations. 

On a negative one, however, three rules were considered incomplete with respect to original guideline content. 
Moreover, while four rules were devoted to the diagnosis of subarachnoid hemorrhage, the diagnosis of cerebral hemorrhage was not treated in depth. 

In order to solve these issues, as already observed, we have implemented a human-in-the-loop strategy, where the physician identified the problems, and (with the help of a technical expert) refined the prompt. In the case of cerebral hemorrhage rules, for instance, the prompt instructed the LLM to focus on such a diagnosis. In the next iteration, the LLM was thus asked to provide a more complete/correct rule set. In our experiments, one further iteration was enough.  


After filtering, only 50 rules survived, as applicable to the event log; as we already observed, most traces deal just with the first four rule categories (since other steps, such as prevention, do not take place in a neurological ward), which justifies most of the exclusions. In other cases, rules were not applicable because the data needed for the check were not collected on the patients at hand, or were not reported in the discharge letter. In the future, we may consider the opportunity to integrate the discharge letter information with other hospital data sources, for the sake of completeness.

As an example, we show a rule which deals with the acute phase management of ischemic stroke: \emph{IF a patient has an occlusion of the intracranial internal carotid artery, middle cerebral artery,
M1-M2 segments, or anterior cerebral artery, A1 segment, AND is unresponsive or unable to undergo
intravenous thrombolysis, THEN mechanical thrombectomy techniques are recommended within 6
hours of symptom onset.}

Figure \ref{python_Rule30} reports part of the corresponding Python code, as produced after the refinement phase. As it can be observed, the script takes advantage of the activity synonym reference list to facilitate conformance checking of patient traces (where synonyms of the medical terms might have been used - see, e.g., the $thrombectomy\_terms$ list in figure  \ref{python_Rule30}). Moreover, temporal constraints are properly calculated.

\begin{figure*}
\begin{adjustbox}{max totalsize={1.4\textwidth}{0.98\textheight},varwidth=\linewidth}
\begin{lstlisting}[language=Python,frame=lines,numbers=left]
def rule_thrombectomy(trace: list[dict]) -> str:
    def _search_term(event: dict, terms: list) -> bool:
        for k, v in event.items():
            val = str(v).lower()
            for t in terms:
                if t.lower() in val: 
                    return True
        return False

    occlusion_ica_terms = ["Occlusion of the intracranial internal carotid artery",
                           "Intracranial ICA near-occlusion" , ...]
    thrombectomy_terms = ["Mechanical thrombectomy", "Endovascular thrombectomy", ... ]
    onset_terms = ["stroke onset", "onset", "stroke detection", ... ]
    ...

    for event in trace:
        if not has_occlusion_ica and _search_term(event, occlusion_ica_terms):
            has_occlusion_ica = True
            ...
        if onset_ts is None and _search_term(event, onset_terms):
            current_ts = parse_timestamp(event.get('time:timestamp'))
            if current_ts:
                onset_ts = current_ts
        if _search_term(event, trombolysis_iv_terms):
            iv_trombolysis_administered = True
            ...
    if onset_ts is None and len(trace) > 0:
        onset_ts = parse_timestamp(trace[0].get('time:timestamp'))
    ...
    
    if onset_ts is None:
        return "NOT APPLICABLE"

    if mechanical_thrombectomy_performed:
        time_delta_minutes = get_time_delta_minutes(onset_ts, mechanical_thrombectomy_ts)

        if time_delta_minutes is not None and time_delta_minutes <= 360:
            return "CONFORMANT"
        else:
            return "NOT CONFORMANT"
    else:
        return "NOT CONFORMANT" 

\end{lstlisting}
\end{adjustbox}
\caption{Part of the Python code generated by Gemini  for an example rule. In the figure, only some sections of the code are shown, for the sake of simplicity, and terms in the activity synonym reference list have been translated into English}
\label{python_Rule30}
\end{figure*}

\begin{figure*}
\includegraphics[width=\textwidth]
{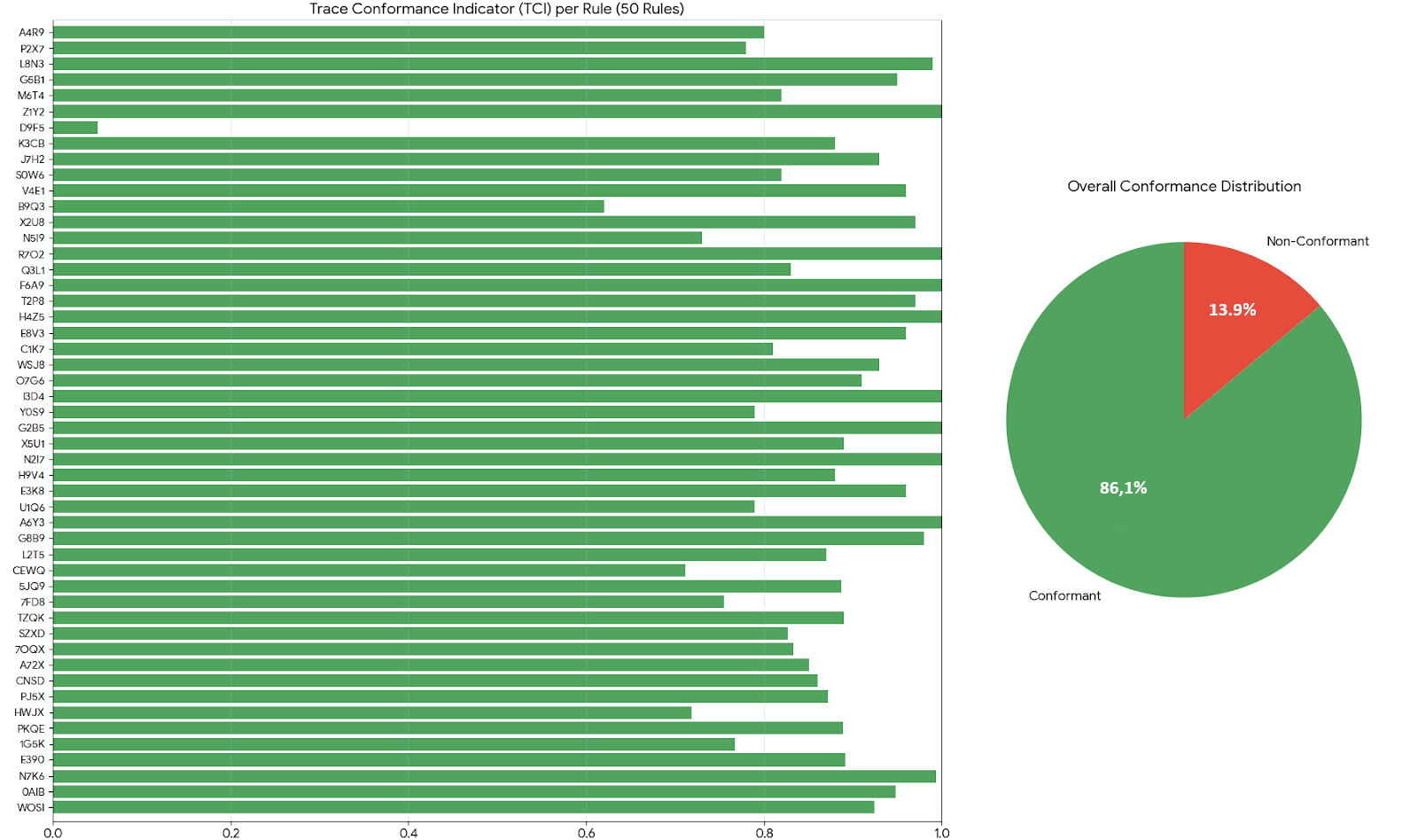}
\caption{Conformance analysis results.}
\label{risu-marzone}
\end{figure*}

Figure \ref{risu-marzone} shows a summary of our experimental results on conformance checking at the neurological ward of Alessandria Hospital. 

On the 50 rules, more than 86\% were conformant to the clinical guideline, testifying a robust quality assessment performance for Alessandria. 
Nevertheless, the (few) non conformance examples testify that there is some space for further improvements. In the following, as an example, we report our investigation on two specific rules, one with a medium TCI value, and the other with a very low one. 

The first rule is expressed as follows: "IF a patient has cerebral hemorrhage, THEN it is suggested to start treating their high blood pressure and reversing any anticoagulant therapies as soon as they arrive at the emergency room". 
Such a rule obtained a TCI value of 62\%. In most cases, non conformance was due to a delay to therapy administration, which was managed only in the neurological ward, and not in the emergency room, testifying a suboptimal communication between the two wards. In a specific example, however, looking at the trace and at the corresponding discharge letter, we were able to identify a delayed diagnosis: the patient was initially treated as the victim of a home accident, that caused the haemorrage, and only later it was understood that an ischemic stroke had determined the accident itself, depicting a situation which was more complex than a simple cooordination bottleneck. 

We also investigated a second rule, expressed as follows: "If a patient has ischemic stroke AND hyperthermia, THEN pharmacological correction of hyperthermia is indicated (preferably with paracetamol), maintaining the temperature below 37°C". This rule totalized a TCI value of only 6\%. After analyzing the not conformant traces, we found that all the patients involved suffered from hyperthermia, but their treatment was different from paracetamol. Interestingly, however, most of these patients underwent a diagnostic path, which allowed physicians to identify the precise cause of hyperthermia, leading to a different, more proper therapeutic strategy. In particular, 69\% of the traces are related to patients who had contracted bacterial infections, and were thus treated with antibiotics; 6\% of the traces relate to Covid-19-infected patient, who were treated with antivirals and anti-inflammatories. While these examples technically represent non conformances, in fact they were correct adaptations of the therapeutic strategy to the specific situations, with the objective of tackling the cause of the problem, instead of just reducing the symptoms (as paracetamol would do). The few remaining cases refer to patients having non-problematic hyperthermia (brief episodes resolved spontaneously), or to patients developing hyperthermia just before discharge or transfer to other wards, probably treated at destination.


In conclusion, our tool enabled us to identify and examine the non conformance examples experienced in Alessandria, highlighting some existing issues (e.g., a reduced coordination between the emergency room and the neurological ward), but also sheding light on cases which are only apparently suboptimal.

\section{Conclusion}
\label{conclu}

In this paper, we have proposed an architecture that integrates multiple LLMs and other supporting components to: (1) extract patient traces from clinical discharge letters, (2) infer normative rules from a textual clinical guideline,  (3) formalize the relevant rules into executable representations, and (4) compute the $\mathcal{TCI}$ indicator, to measure trace conformance, supporting quality assessment without requiring a time-consuming formalization of the guideline itself.

Our contribution relies on the recent AI trend that moves towards system-level orchestration design, and, to the best of our knowledge, it is the first one able to cover the four tasks listed above, in an automatic and integrated way. 

Our experiments within the stroke management domain have showcased the real-world usability of the approach, while at the same time verifying the globally high quality of stroke care at the neurological ward of Alessandria Hospital, where major issues  were not identified, but a possible organizational bottleneck was found.

Collaboration with medical specialists, who verified the consistency of the extracted traces and rules with discharge letters and with established medical knowledge, has ensured the semantic validity of the current version of our approach.

Nevertheless, we consider it valuable to evaluate trace/rule extraction testing alternative LLMs, particularly models trained on healthcare-specific data, such as Google’s Med-Gemini \citep{yang2024advancingmultimodalmedicalcapabilities}, which might derive even clearer and accurate information, potentially reducing the effort required during expert validation.

We also aim to advance toward the (partial) automation of the trace/rule verification process.  To this end, we plan to explore the research paradigm known as LLM-as-a-Judge \citep{Gu}, in which LLMs are used to evaluate the outputs of other LLMs on complex tasks. Although human assessment remains the gold standard, it is often time-consuming and expensive; in contrast, LLM-based evaluation can provide faster feedback and more consistent criteria, assuming that a well-defined evaluation framework is in place. While the adoption of LLM-as-a-Judge methods must be approached with caution in medical contexts, such techniques could nonetheless accelerate validation by assigning an initial quality score, with selected cases subsequently reviewed by human experts when necessary. 


Additionally, in the future we plan to rely on the current framework to develop an education facility, testing the capability of non-expert physicians in learning the guideline and identifying non-conformant traces with respect to it, and supporting them towards the acquisition of a stronger competence in the field.

\noindent {\bf Ethics Committee Approval}
The study was approved by Comitato Etico Territoriale  Interaziendale AOU Maggiore della Carità di Novara (protocol code 216/CE and date of approval 19 March 2025).





\printcredits

\end{document}